\documentclass[letterpaper]{article} 
\usepackage{aaai25}  
\usepackage{times}  
\usepackage{helvet}  
\usepackage{courier}  
\usepackage[hyphens]{url}  
\usepackage{graphicx} 
\usepackage{natbib}  
\usepackage{caption} 
\usepackage{algorithm}
\usepackage{algorithmic}

\usepackage{newfloat}
\usepackage{listings}
\DeclareCaptionStyle{ruled}{labelfont=normalfont,labelsep=colon,strut=off} 
\floatstyle{ruled}
\newfloat{listing}{tb}{lst}{}
\floatname{listing}{Listing}
\usepackage{bm}
\usepackage{enumitem}
\usepackage{multirow}
\usepackage{graphicx}
\usepackage{subfigure}
\usepackage{bbding}
\usepackage{pifont}
\usepackage{color}
\usepackage{booktabs}
\usepackage{amsmath}

\newcommand{\model}{\textsc{MyGO}}

\title{Tokenization, Fusion, and Augmentation: Towards Fine-grained \\ Multi-modal Entity Representation}
\author{
    Yichi Zhang\textsuperscript{\rm 1,2}, Zhuo Chen\textsuperscript{\rm 1,2}, Lingbing Guo\textsuperscript{\rm 1,2}, Yajing Xu\textsuperscript{\rm 1,2}, Binbin Hu\textsuperscript{\rm 3}, Ziqi Liu\textsuperscript{\rm 3}\\Wen Zhang\textsuperscript{{\rm 4,2}}\footnotemark[1], Huajun Chen\textsuperscript{\rm 1,2,5}\footnote{Corresponding Authors.}
}
\affiliations{
    \textsuperscript{\rm 1}College of Computer Science and Technology, Zhejiang University\\
    \textsuperscript{\rm 2}ZJU-Ant Group Joint Lab of Knowledge Graph\\
    \textsuperscript{\rm 3}Ant Group\\
    \textsuperscript{\rm 4}School of Software Technology, Zhejiang University\\
    \textsuperscript{\rm 5}Zhejiang Key Laboratory of Big Data Intelligent Computing\\

    \{zhangyichi2022, zhuo.chen, zhang.wen, huajunsir\}@zju.edu.cn
}

\usepackage{bibentry}

\begin{document}

\maketitle

\begin{abstract}
Multi-modal knowledge graph completion (MMKGC) aims to discover unobserved knowledge from given knowledge graphs, collaboratively leveraging structural information from the triples and multi-modal information of the entities to overcome the inherent incompleteness. Existing MMKGC methods usually extract multi-modal features with pre-trained models, resulting in coarse handling of multi-modal entity information, overlooking the nuanced, fine-grained semantic details and their complex interactions. To tackle this shortfall, we introduce a novel framework {\model} to \textbf{tokenize, fuse, and augment the fine-grained multi-modal representations of entities} and enhance the MMKGC performance. Motivated by the tokenization technology, {\model} tokenizes multi-modal entity information as fine-grained discrete tokens and learns entity representations with a cross-modal entity encoder. To further augment the multi-modal representations, {\model} incorporates fine-grained contrastive learning to highlight the specificity of the entity representations. Experiments on standard MMKGC benchmarks reveal that our method surpasses 19 of the latest models, underlining its superior performance. Code and data can be found in \url{https://github.com/zjukg/MyGO}.
\end{abstract}

%

\section{Introduction}
\textbf{Multi-modal knowledge graphs (MMKGs)} \cite{MMKG-Survey} encapsulate diverse and complex world knowledge as structured triples (\textit{head entity, relation, tail entity}) while incorporating multi-modal data such as images and text for additional entity context. These extensive triples, alongside their multi-modal content, form a vast multi-modal semantic network that constitutes significant infrastructures for many fields, such as recommendation \cite{MMKG-rec}, multi-modal understanding \cite{MMKG-pretraining}, and large language models \cite{MMKG-LLM}. MMKGs furnish these systems with a dependable source of factual knowledge.
\par MMKGs frequently grapple with the challenge of incompleteness as considerable amounts of valid knowledge remain undiscovered during their creation. This phenomenon underscores the importance of \textbf{multi-modal knowledge graph completion (MMKGC)} \cite{MMKG-Survey}, which aims to identify new knowledge from the given MMKGs automatically. Unlike conventional knowledge graph completion (KGC) \cite{liang2024mines} that predominantly focuses on modeling the triple structure based on the existing KGs, MMKGC needs to manage the additional multi-modal information that enriches entity description from various perspectives. Therefore, the essence of MMKGC is to harmoniously integrate structural information from triples with the rich multi-modal features associated with entities. This synergy is pivotal for informed knowledge inference within the embedding space, where the rich multi-modal information of entities serves as supplementary information and provides robust and effective multi-modal features for MMKGC.

\begin{figure}
  \centering
  {\includegraphics[width=0.9\columnwidth]{pictures/introduction.pdf}
  }
  \caption{An intuition of existing MMKGC methods and {\model}. {\model} attempts to tokenize raw modality data into fine-grained tokens and learn the fine-grained entity representations to model semantic unit interactions.}
  \label{figure:introduction}
\end{figure}

\par Existing MMKGC methods \cite{sergieh_multimodal_2018-TBKGC} tend to represent modality information as single embedding derived from pre-trained models \cite{BERT}, utilizing a fusion and prediction module to measure the triple plausibility. However, this paradigm is rather simplistic and frequently fails to capture the intricate details in the modality data. Typically, in this paradigm, the modality information extracted by pre-trained models would be frozen in later training. Moreover, when handling multiple modality instances, such as several images of an entity, these methods resort to vanilla operations like averaging, thereby stripping away potentially significant details.
Considering the raw modality data houses detailed semantic units to present the crucial entity features, the common practice of generating a static embedding per modality can lead to a loss of valuable granular information, subsequently restricting MMKGC model performance. For instance, we present a simple case in Figure \ref{figure:introduction} to show these fine-grained semantic units in the image and text of an entity T-Rex, which are the image segments and the textual phrases. These fine-grained semantic features not only describe an entity but also embody complex cross-modal relationships. We advocate for a more fine-grained framework, allowing MMKGC models to capture the subtle, shared information embedded within the data through detailed interactions. This approach promises to significantly augment entity representations.

\par Aiming to solve the fine-grained information processing and leveraging problem, we propose a novel framework {\model} to achieve \textbf{fine-grained multi-modal information processing, interaction, and augmentation} in MMKGC models. Figure \ref{figure:introduction} gives a clear contrast between existing MMKGC methods and our {\model}. {\model} first employs a \textbf{modality tokenization (MT)} module to tokenize the entity modality information in MMKGs into fine-grained discrete token sequences using existing pre-trained tokenizers \cite{BEIT}, followed by learning the MMKGC task through a \textbf{hierarchical triple modeling (HTM)} architecture. HTM consists of a cross-modal entity encoder, a contextual triple encoder, and a relational decoder to encode the fine-grained entity representation and measure the triple plausibility. To further augment and refine the entity representations, we propose a \textbf{fine-grained contrastive loss (FGCL)} to generate varied contrastive samples and boost the model performance. We conduct comprehensive experiments with public MMKG benchmarks \cite{MMKG}. Comparisons against 19 recent baselines demonstrate the outperforming results of {\model}. We also delve further into the nuances of {\model}'s design to understand it. Our contribution is three-fold:
\par (1). We emphasize fine-grained multi-modal learning for MMKGC and propose a cutting-edge framework {\model}. {\model} tokenizes the modality data into fine-grained multi-modal tokens and pioneers a novel MMKGC architecture to hierarchically model the cross-modal entity representation.
\par (2). We propose a fine-grained contrastive learning module to augment the cross-modal entity representations. This module innovates by employing new tactics to generate high-quality comparative samples for more detailed and effective self-supervised contrastive learning.
\par (3). We conduct comprehensive experiments on public benchmarks and achieve new state-of-the-art performance against 19 baselines with further exploration.

\section{Related Works}
\begin{figure*}[]
  \centering
\includegraphics[width=0.85\linewidth]{pictures/model.pdf}
  \caption{The overview of our {\model} framework. We mainly have three parts of new designs in {\model} to tokenize, fuse, and augment the fine-grained multi-modal semantic information in the MMKGs.} 
  \label{figure::model}
\end{figure*}
Multi-modal knowledge graphs (MMKGs) \cite{MMKG-Survey, KGSurvey} are knowledge graphs with rich multi-modal information like images, text descriptions, audio, and videos \cite{DBLP:conf/mm/WangMCML023-TIVA}. Due to the incompleteness of the knowledge graphs, knowledge graph completion (KGC) \cite{contrastive} is a popular research topic to automatically discover unobserved knowledge triples by learning from the triple structure. \textbf{Multi-modal knowledge graph completion (MMKGC)} \cite{SPTformer} aims to predict missing triples in the given MMKGs collaboratively leveraging the extra multi-modal information from entities. Existing MMKGC methods mainly make new improvements in three perspectives: (1) multi-modal fusion and interaction \cite{cao_otkge_2022-OTKGE}, (2) integrated decision \cite{li_imf_2023-IMF}, and (3) negative sampling \cite{MMRNS}.

\section{Task Definition}
\label{task}
A multi-modal knowledge graph (MMKG) incorporating both visual and textual modalities can be represented as $\mathcal{G}=(\mathcal{E}, \mathcal{R}, \mathcal{T}, \mathcal{V}, \mathcal{D})$, where $\mathcal{E}, \mathcal{R}$ are the entity set and the relation set. $\mathcal{T} = \{(h, r, t)\mid h, r\in\mathcal{E}, r\in\mathcal{R}\}$ is the triple set, indicating that entity $h$ is related to entity  through $t$ relation $r$. Besides, $\mathcal{V}, \mathcal{D}$ correspond to the collections of images and textual descriptions for each entity $e$.
\par The primary aim of \textbf{knowledge graph completion (KGC)} is to learn a score function $\mathcal{S}(h, r, t)$ which measures the plausibility of a triple $(h, r, t)$ by \textbf{a scalar score}. In KGC models, entities and relations correspond to embeddings, and the triple score is defined on these embeddings, preferring high scores for positive triples and lower scores for negative triples. In other words, the plausibility of positive triples in the training set is maximized by a positive-negative contrast \cite{bordes_translating_2013-TransE} during training. Expanding to MMKGs, \textbf{multi-modal knowledge graph completion (MMKGC)} would further consider the multi-modal information $\mathcal{V}(e), \mathcal{D}(e)$ to enhance their embeddings.

\section{Methodology}
In this section, we will detailedly introduce the framework proposed by us which leverages \textbf{\underline{M}}odalit\textbf{\underline{Y}} information as fine-\textbf{\underline{G}}rained t\textbf{\underline{O}}kens ({\model} for short) to \textbf{tokenize, fuse, and augment the fine-grained multi-modal representations of entities}. We consider the mainstream MMKGC setting \cite{MMRNS} that includes both image and text modalities \cite{MMKG-Survey}. {\model} mainly comprises three modules: modality tokenization module, hierarchical triple modeling module, and fine-grained contrastive learning, aiming to process, fuse, and augment the fine-grained information in MMKGs respectively. Figure \ref{figure::model} provides an intuitive perspective of the design of {\model}.

\subsection{Modality Tokenization}
\label{section::tokenization}
To capture fine-grained multi-modal information, we propose a \textbf{modality tokenization (MT)} module to process the raw multi-modal data of entities into fine-grained discrete semantic tokens, serving as semantic units to learn fine-grained entity representations. We employ the tokenizers for image and textual modality respectively, denoted as $\mathcal{Q}_{img}, \mathcal{Q}_{txt}$, to generate visual tokens $v_{e, i}$ and textual tokens $w_{e, i}$ for entity $e$:
\begin{equation}
    \mathcal{U}_{img}(e) = \{v_{e, 1}, v_{e, 2},\cdots, v_{e, m_e}\} = \mathcal{Q}_{img}(\mathcal{V}(e))
\end{equation}
\begin{equation}
    \mathcal{U}_{txt}(e) = \{w_{e, 1}, w_{e, 2},\cdots, w_{e, n_e}\} = \mathcal{Q}_{txt}(\mathcal{D}(e))
\end{equation}
where $m_e, n_e$ is the number of tokens for each modality and we denote $\mathcal{U}(e)$ symbolizes the collective token set for entity $e$. The text tokens are from the vocabulary of a language model \cite{BERT} while the visual tokens are from the codebook of a pre-trained visual tokenizer \cite{VQ-GAN, BEIT}. Notably, $\mathcal{V}(e)$ might consist of multiple images and we process each image to accumulate the tokens in $\mathcal{U}_{img}(e)$. 

\par During the tokenization process, it's common to encounter duplicate tokens. Therefore, we count the occurrence frequency of each token, retaining a predetermined quantity of the most common tokens for each modality. Additionally, we remove the stop words \cite{stopwords} in the textual descriptions as their contribution to the entity semantics is minimal. After the MT and refinement process, we can obtain processed token set $\mathcal{U}_{img}'$ and $\mathcal{U}_{txt}'$ with $m$ visual tokens and $n$ textual tokens for each entity $e$, featuring a collection of fine-grained tokens that embody the vital features derived from the raw multi-modal data. Subsequently, we assign a separate embedding for each token in $\mathcal{U}_{img}'$ and $\mathcal{U}_{txt}'$. This approach is tailored to the fact that different entities may share tokens, and individualized embeddings of tokens allow for a more fine-grained representation of similar features across various entities, enriching the entities' profiles with detailed multi-modal semantic units.

\subsection{Hierarchical Triple Modeling}
After the MT process, we further design a \textbf{hierarchical triple modeling (HTM)} module in this section. HTM leverages a hierarchical transformer architecture to capture the multi-modal entity representation and model the triple plausibility in a hierarchical manner, which consists of three components: cross-modal entity encoder, contextual triple encoder, and relational decoder.

\subsubsection{\textup{\textbf{Cross-modal Entity Encoder}}}
The cross-modal entity encoder (CMEE) aims to capture the multi-modal representation of entities by leveraging their fine-grained multi-modal tokens. Unlike existing methods \cite{li_imf_2023-IMF}, {\model} performs fine-grained tokenization and obtains a sequence of discrete tokens. Therefore, we design a more \textbf{fine-grained feature interaction method} that allows full interaction between all the different modal messages. In {\model}, we apply a transformer \cite{transformer} layer as the CMEE. We first linearize the multi-modal tokens as a sequence:
\begin{equation}
    \mathcal{X}(e)=([\mathtt{ENT}], s_e, v_{e, 1},\cdots, v_{e, m}, w_{e, 1},\cdots, w_{e, n})
\end{equation}
where $[\mathtt{ENT}]$ is a special token and $s_e$ is a learnable embedding representing the structural information of the entity. $[\mathtt{ENT}]$ is analogous the $[\mathtt{CLS}]$ token in BERT \cite{BERT} to capture the sequence feature for downstream prediction. $s_e$ is a learnable embedding to represent the structural information learned from the existing triple structures, which will be optimized during training. 
Besides, for the multi-modal tokens from $\mathcal{U}_{img}'$ and $\mathcal{U}_{txt}'$, we freeze their initial representations derived from the tokenizers and define linear projection layers $\mathcal{P}_{img},\mathcal{P}_{txt}$ to project them into the same representation space as $
\widehat{v}_{e, i}=\mathcal{P}_{img}({v}_{e, i})+b_{img}\quad\widehat{w}_{e, j}=\mathcal{P}_{txt}({v}_{e, j})+b_{txt}
$ where $b_{img}, b_{txt}$ are defined modality biases to enhance the labeling of information from distinct modalities.
In this way, the final sequence entered into CMEE becomes ${\mathcal{X}_{input}(e)}=([\mathtt{ENT}], s_e, \widehat{v}_{e, 1},\cdots, \widehat{v}_{e, m}, \widehat{w}_{e, 1},\cdots, \widehat{w}_{e, n})$. 
The cross-modal entity representation is obtained by $\mathbf{e}=\mathbf{Pooling}(\mathbf{Transformer}(\mathcal{X}_{input}(e)))$,
where $\mathbf{Transformer}()$ represents a transformer encoder layer \cite{transformer}, $\mathbf{Pooling}$ is the pooling operation with obtains the final hidden representation of the special token $[\mathtt{ENT}]$. It allows each token in the input sequence can be dynamically highlighted by CMEE to interact and eventually learn expressive entity representations.

\subsubsection{\textup{\textbf{Contextual Triple Encoder}}}
To achieve adequate modality interaction in the relational context, we apply another transformer layer as the \textbf{contextual triple encoder (CTE)} to encode the contextual embeddings for the given query. Taking head query $(h, r, ?)$ (tail prediction) as an example, we can obtain the contextual embeddings $\tilde{\mathbf{h}}$ as:
$\tilde{\mathbf{h}} = \mathbf{Transformer}([\mathtt{CXT}], \mathbf{h}, \mathbf{r})$,
where $[\mathtt{CXT}]$ is a special token in the input sequence to capture the contextual embedding of entity, $\mathbf{h}$ is the output representation of $h$ from CMEE, and $\mathbf{r}$ is the relation embedding for each $r\in\mathcal{R}$. The contextual embeddings of the query $(h, r, ?)$ are then processed by a relational decoder for entity prediction.

\subsubsection{\textup{\textbf{Relational Decoder}}}
Moreover, we employ a score function $\mathcal{S}(h, r, t)$ to measure the triple plausibility by producing a scalar score, which functions as the relational decoder for query prediction. In {\model}, we employ Tucker \cite{Tucker} as our score function denoted as $\mathcal{S}(h, r, t) = \mathcal{W}\times_1\tilde{\mathbf{h}}\times_2\tilde{\mathbf{r}}\times_3\mathbf{t}$,
where $\times_i$ represents the tensor product along the i-th mode, $\mathcal{W}$ is the core tensor learned during training. We train our model with cross-entropy loss for each triple. We treat $t$ as the golden label against the whole entity set $\mathcal{E}$, which is the same for head prediction.  Therefore, the objective is a cross-entropy loss:
\begin{equation}
    \mathcal{L}_{head}=-\sum_{(h, r, t)\in\mathcal{T}}\log\frac{\exp(\mathcal{S}(h, r, t))}{\sum_{t'\in\mathcal{E}}\exp(\mathcal{S}(h, r, t'))}
\end{equation}
Note that we use the contextual embedding $\tilde{\mathbf{e}}_{h}$ of $h$ and the multi-modal embedding $\mathbf{e}_{t}$ of $t$ to calculate the score, which can expedite computation. Otherwise, we would need to extract the contextual embedding of all the candidate entities under different relations, which needs $O(|\mathcal{E}|\times|\mathcal{R}|)$-level forward passes in contextual transformer and would greatly increase the computation of the model.
Besides, both head prediction and tail prediction are considered in {\model}, and the objective $\mathcal{L}_{tail}$ is similar when giving a tail query $(?, r, t)$. The overall MMKGC task objective can be denoted as $\mathcal{L}_{kgc}=\mathcal{L}_{head}+\mathcal{L}_{tail}$.

\subsection{Fine-grained Contrastive Learning}
Based on the above design, we have been able to train and test the MMKGC model. To further augment fine-grained and robust multi-modal entity representations, we introduce a \textbf{fine-grained contrastive learning (FGCL)} module in {\model} to achieve this goal by multi-scale contrastive learning on the entity representations.

\par As mentioned before, CMEE aims to capture the entity representation based on a multi-modal token sequence. Inspired by the idea of SimCSE \cite{SIMCSE}, we augment these entity representations through contrastive learning. Specifically, given an entity $e$, we can get two representations $\mathbf{e}, \mathbf{e}_{sec}$ from CMEE by two forward passes. The variations between these two embeddings, induced by the dropout layer in the transformer encoder, allow for slight deactivation of multi-modal token features, effectively acting as a form of simple data augmentation. By in-batch contrastive learning across a collection of entities, {\model} is trained to extract truly significant information from token sequences, thereby enhancing the distinctiveness of each entity's representation. To deepen the granularity of this process, we further extract three additional representations from the transformer output, which can represent entity features from their perspectives. We can define the output representations of the multi-modal tokens in an input sequence ${\mathcal{X}_{input}(e)}$ as:
${\mathcal{X}_{output}(e)}=([\mathtt{ENT}]', s_e', \widehat{v}_{e, 1}',\cdots, \widehat{v}_{e, m}', \widehat{w}_{e, 1}',\cdots, \widehat{w}_{e, n}')$
Then we introduce three embeddings $\mathbf{s}(e), \mathbf{v}(e), \mathbf{w}(e)$ to represent the global, visual, and textual information of an entity $e$. $\mathbf{s}(e)$ is derived from the average of all the output representations in $\mathcal{X}_{output}(e)$. Similarly, $\mathbf{v}(e)$ and $\mathbf{w}(e)$ are the averages of the corresponding visual and textual tokens. They can be denoted as:
\begin{equation}
    \begin{aligned}
            \mathbf{s}(e)=&\mathbf{Mean}(\mathcal{X}_{output}(e))\\
            \mathbf{v}(e)=\frac{1}{m}\sum_{i=1}^{m}&\widehat{v}_{e, i}' \quad
    \mathbf{w}(e)=\frac{1}{n}\sum_{i=1}^{n}\widehat{w}_{e, i}'
    \end{aligned}
\end{equation}
Among these embeddings, $\mathbf{e}_{sec}$, $\mathbf{s}(e)$ encapsulate the global information of $e$ and $\mathbf{v}(e),\mathbf{w}(e)$  consist of the local modality information of the entity $e$. For each entity $e$, we can collect its candidates for contrastive learning as $\mathcal{C}(e)=\{e_{sec}, \mathbf{s}(e), \mathbf{v}(e), \mathbf{w}(e)\}$, which consists of its global and local features. $(\mathbf{e}, \mathbf{e}')$ where $\mathbf{e}'\in\mathcal{C}(e)$ is regarded as a positve sample. Then we employ in-batch negative sampling to construct negative pairs and the final objective can be denoted as:
\begin{equation}
    \mathcal{L}_{con}=-\sum_{i=1}^{\mathcal{B}}\sum_{e'_i\in\mathcal{C}(e_i)}\log\frac{\exp(\mathbf{cos}(\mathbf{e}_i, \mathbf{e}_i')/\tau)}{\sum_{j=1}^{\mathcal{B}}\exp(\mathbf{cos}(\mathbf{e}_i, \mathbf{e}_j')/\tau)}
\end{equation}
where $\mathcal{B}$ is the batch size, $\mathbf{cos}(\cdot,\cdot)$ is the cosine similarity of two embeddings and $\tau$ is the temperature hyper-parameter. Through such an FGCL process, {\model} notably improves its ability to discern detailed multi-modal attributes across various entities, boosting the model performance in the MMKGC task. Finally, the overall training objective of our framework can be denoted as $\mathcal{L}=\mathcal{L}_{kgc}+\lambda
\mathcal{L}_{con}$ where $\lambda$ is a hyper-parameter to control the weight of $\mathcal{L}_{con}$.




\section{Experiments}
In this section, we will conduct comprehensive experiments to evaluate the performance of {\model}. We begin by detailing our experimental setup and subsequently present the \textbf{effectiveness}, \textbf{reasonablity}, \textbf{efficiency} and \textbf{explainability} of {\model}. 

\begin{table*}[t]

\centering
\setlength{\tabcolsep}{0.5mm}{
\begin{tabular}{c|c|cccc|cccc|cccc}
\toprule
\multicolumn{1}{c|}{\multirow{2}{*}{\textbf{Model}}} & \multirow{2}{*}{\textbf{\begin{tabular}[c]{@{}c@{}}Fusion\\ Strategy\end{tabular}}} & \multicolumn{4}{c|}{\textbf{DB15K}} & \multicolumn{4}{c|}{\textbf{MKG-W}} & \multicolumn{4}{c}{\textbf{MKG-Y}}\\
 &  & \textbf{MRR} & \textbf{H@1} & \textbf{H@3} & \textbf{H@10} & \textbf{MRR} & \textbf{H@1} & \textbf{H@3} & \textbf{H@10} & \textbf{MRR} & \textbf{H@1} & \textbf{H@3} & \textbf{H@10}\\
\bottomrule
\textbf{TransE} & None & 24.86 & 12.78 & 31.48 & 47.07 & 29.19 & 21.06 & 33.20 & 44.23 & 30.73 & 23.45 & 35.18 & 43.37\\
\textbf{DistMult} & None & 23.03 & 14.78 & 26.28 & 39.59 & 20.99 & 15.93 & 22.28 & 30.86& 25.04 & 19.33 & 27.80 & 35.95\\
\textbf{ComplEx}  & None & 27.48 & 18.37 & 31.57 & 45.37 & 24.93 & 19.09 & 26.69 & 36.73& 28.71 & 22.26 & 32.12 & 40.93\\
\textbf{RotatE} & None & 29.28 & 17.87 & 36.12 & 49.66 & 33.67 & 26.80 & 36.68 & 46.73& 34.95 & 29.10 & 38.35 & 45.30\\
\textbf{TuckER} & None &  \underline{33.86} & \underline{25.33} & 37.91 & 50.38 & 30.39 & 24.44 & 32.91 & 41.25 & 37.05 & \underline{34.59} & 38.43 & 41.45
\\
\midrule
\textbf{IKRL} & Static & 26.82 & 14.09 & 34.93 & 49.09 & 32.36 & 26.11 & 34.75 & 44.07& 33.22 & 30.37 & 34.28 & 38.26
\\
\textbf{TBKGC}  & Static & 28.40 & 15.61 & 37.03 & 49.86 & 31.48 & 25.31 & 33.98 & 43.24& 33.99 & 30.47 & 35.27 & 40.07
\\
\textbf{TransAE} & Static & 28.09 & 21.25 & 31.17 & 41.17 & 30.00 & 21.23 & 34.91 & 44.72& 28.10 & 25.31 & 29.10 & 33.03
\\
\textbf{MMKRL}  & Static & 26.81 & 13.85 & 35.07 & 49.39 & 30.10 & 22.16 & 34.09 & 44.69& 36.81 & 31.66 & 39.79 & 45.31
\\
\textbf{RSME} & Adaptive & 29.76 & 24.15 & 32.12 & 40.29 & 29.23 & 23.36 & 31.97 & 40.43 & 34.44 & 31.78 & 36.07 & 39.09
\\
\textbf{VBKGC} & Static & 30.61 & 19.75 & 37.18 & 49.44 & 30.61 & 24.91 & 33.01 & 40.88& 37.04 & 33.76 & 38.75 & 42.30
\\
\textbf{OTKGE} & Adaptive & 23.86 & 18.45 & 25.89 & 34.23 & 34.36 & \underline{28.85} & 36.25 & 44.88& 35.51 & 31.97 & 37.18 & 41.38
\\
\textbf{MACO} & Static & 27.41 & 14.61 & 35.59  & 50.00 & 31.74 & 25.23 & 34.23  & 44.37 & 34.98 & 31.59 & 36.68 & 40.51 \\
\textbf{IMF} & Adaptive & 32.25 & {24.20} & 36.00 & 48.19 & 34.50 & 28.77 & 36.62 & 45.44& 35.79 & 32.95 & 37.14 & 40.63
\\
\textbf{QEB} & Static & 28.18 & 14.82 & 36.67 & 51.55 & 32.38 & 25.47 & 35.06 & 45.32& 34.37 & 29.49 & 36.95 & 42.32
\\
\textbf{VISTA}& Adaptive & 30.42 & 22.49 & 33.56 & 45.94 & 32.91 & 26.12 & 35.38 & 45.61& 30.45 & 24.87 & 32.39 & 41.53
\\
\textbf{AdaMF} & Adaptive & 32.51 & 21.31 & \underline{39.67} & \underline{51.68} & 34.27 & 27.21 & \underline{37.86} & 47.21& \underline{38.06} & 33.49 & \textbf{40.44} & \textbf{45.48}
\\

\midrule
\textbf{MANS} & Static & 28.82 & 16.87 & 36.58 & 49.26 & 30.88 & 24.89 & 33.63 & 41.78& 29.03 & 25.25 & 31.35 & 34.49
\\
\textbf{MMRNS} & Adaptive & 32.68 & 23.01 & 37.86 & 51.01 & \underline{35.03} & 28.59 & 37.49 & \underline{47.47}& 35.93 & 30.53 & 39.07 & \underline{45.47}
\\
\midrule
\multicolumn{1}{c|}{\multirow{2}{*}{\textbf{\model}}} & \multirow{2}{*}{Adaptive}  & \textbf{37.72} & \textbf{30.08} & \textbf{41.26} & \textbf{52.21} & \textbf{36.10} & \textbf{29.78} & \textbf{38.54} & \textbf{47.75}& \textbf{38.44} & \textbf{35.01} & \underline{39.84} & 44.19
\\
\multicolumn{1}{c|}{} & & +11.4\%  & +18.4\% & +4.0\%  & +1.0\%  & +3.1\%  & +3.2\%  & +1.8\%  & +0.6\% & +0.9\% &+1.2\% & - & -
\\
\bottomrule
\end{tabular}
}
\caption{The main MMKGC results. We list the type of fusion strategy (none / static / adaptive) considered by each method in the table. The best results are marked as \textbf{bold} and the second best results are \underline{underlined}.}
\label{table::main}
\end{table*}

\subsection{Experiment Settings}
\noindent\textbf{Datasets. }In this paper, we employ three public MMKGC benchmarks DB15K \cite{MMKG}, MKG-W and MKG-Y \cite{MMRNS} to evaluate the model performance. The raw data for each modality are obtained from their official release sources.

\noindent\textbf{Evaluation Protocol.} We conduct link prediction \cite{bordes_translating_2013-TransE} task on the datasets, which is the mainstream MMKGC task. Following existing works, we use rank-based metrics \cite{sun_rotate_2019-RotatE}like mean reciprocal rank (MRR) and Hit@K (K=1, 3, 10) (H@K for short) to evaluate the results. Besides, we employ the filter setting \cite{bordes_translating_2013-TransE} in the prediction results to remove the candidate triples existing in the training data for fair comparisons.

\noindent\textbf{Baselines.} To make a comprehensive performance evaluation, we employ 19 different state-of-the-art MMKGC baselines \cite{bordes_translating_2013-TransE, yang_embedding_2015-DistMult, sun_rotate_2019-RotatE, trouillon_complex_2016-ComplEx, xie_image-embodied_2017-IKRL, sergieh_multimodal_2018-TBKGC, cao_otkge_2022-OTKGE, wang_multimodal_2019-TransAE, DBLP:journals/apin/LuWJHL22-MMKRL, wang_is_2021-RSME, DBLP:journals/corr/abs-2209-07084-VBKGC, MACO, li_imf_2023-IMF, DBLP:conf/mm/WangMCML023-TIVA, lee_vista_2023-VISTA, MAT, MMRNS, DBLP:conf/ijcnn/ZhangCZ23-MANS} in our experiments.

\noindent\textbf{Implementation Details.}
In our experiments, we implement {\model} with PyTorch \cite{pytorch}. For modality tokenization, we employ the tokenizer of BEIT \cite{BEIT} and BERT \cite{BERT} as our visual/textual tokenizers. The codebook size of BEIT \cite{BEIT} is 8192 and the vocabulary size of BERT tokenizer is 32000. During training, we set the training epoch to 2000, the batch size to 1024, and the embedding dimension to 256. The max token number $m$ and $n$ are tuned in $\{4, 8, 12\}$ and the weight $\lambda$ is tuned in $\{1, 0.1, 0.01, 0.001\}$. We optimize the model with Adam \cite{DBLP:journals/corr/KingmaB14-Adam} optimizer.

\subsection{Main Results}
\label{sec::mainexp}
The primary experiment results of MMKGC are depicted in Table \ref{table::main}. We list the strategies in which all methods utilize the modal information in addition to the statistical performance metrics. Unimodal approaches do not incorporate multi-modal information of entities, whereas the multi-modal and negative sampling-based KGC methods are categorized as either static or adaptive based on their multi-modal fusion methodology. {\model} employs self-attention in the encoders so that it is an adaptive method as well.
\begin{figure}[]
  \centering
\includegraphics[width=\columnwidth]{pictures/multi_image.pdf}
  \caption{The MMKGC results of {\model} and several baselines with different numbers of entity images.}
  \label{figure::multi-image}
\end{figure}
\noindent\textbf{Outperforming results.} Firstly, we can observe that {\model} outperforms all baseline methods on all evaluation metrics, achieving new state-of-the-art performance on two datasets. The adaptive MMKGC methods, as indicated by the results, generally outperform static ones. Meanwhile, different from other adaptive approaches, {\model} employs more fine-grained feature processing and fusion with modality tokenization and hierarchical triple modeling. As existing methods tend to set only one feature (embedding) for each modality, {\model} obtains more fine-grained features by tokenization of existing raw data and improves the model performance through fine-grained interactive fusion with transformer-based encoders in HTM.

\noindent\textbf{Accurate reasoning ability.} Further, comparing the improvement of each metric horizontally, we can see that {\model}'s improvement for Hit@1 and MRR is significantly higher than that of Hit@10 and other metrics. For example, on DB15K, {\model} achieves an 18.4\% increase on Hit@1 but a 1\% increase on Hit@10. This underscores {\model}'s capability to significantly improve accurate reasoning through its sophisticated design.

\begin{figure}[t]
  \centering
\includegraphics[width=0.65\columnwidth]{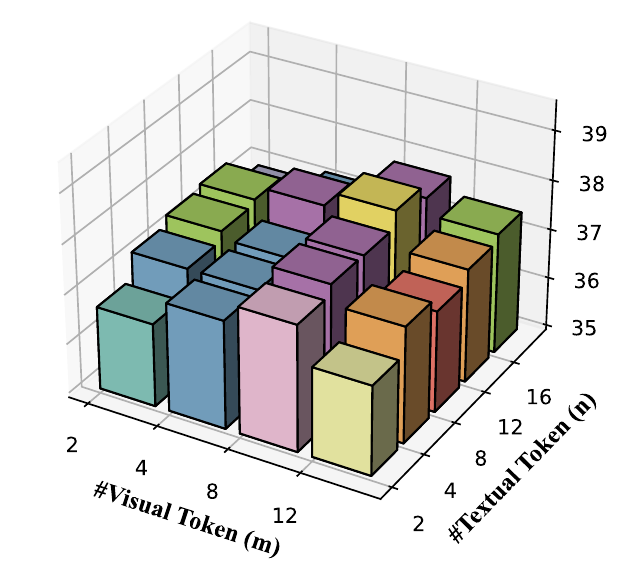}

  \caption{The MRR results with different modality token amount $m$ and $n$ on DB15K.}
  \label{figure::token}
\end{figure}
\subsection{Exploration on Modality Tokenization}
\label{section::subexp}
Compared to current methods, our innovative new design in {\model} introduces the Modality Tokenization (MT). Therefore, we aim to delve deeply into the principles of MT in this section. We mainly focus on two considerable problems: the model's capability to handle multiple pieces of information within a single modality, and the impact of token amounts.
\subsubsection{\textup{\textbf{Multiple Modality Information}}}
Existing methods usually obtain the multi-modal embeddings by consolidating the multiple information within a single modality like mean pooling across various. This operation results in a loss of crucial features and frequently happens during the pre-processing phase. Conversely, MT transforms multiple data particles into a token sequence, preserving common features as much as possible. To demonstrate the effectiveness of MT in this scenario, we conduct another experiment using varying amounts of entity images. As an entity's textual description usually only comprises one paragraph, dividing it is a challenge. Therefore, we evaluate the MMKGC performance of different models on different numbers of entity images, keeping $N=1,2,3,4,5$ images for each entity as far as possible. The results depicted in Figure \ref{figure::multi-image} highlight that in comparison to other baselines, {\model} can achieve consistent and impressive performance enhancements, even when faced with increasing multi-modal data, as the MMKGC performance shows a clear trend of increasing. Contrarily, the baseline performance is somewhat erratic, displaying fluctuations as the image amount increases, and the overall effectiveness does not match {\model}. We attribute this phenomenon to the fact that different methods handle modal information differently. Current models typically generate an embedding for a specific modality from an entity's multiple raw data, thus losing essential original information from the initial features. However, {\model} processes the information into fine-grained semantic units through the design of modality tokenization, retaining the most recurring components. Through this technique, \textbf{{\model} masters the ability to retain the general and uniform information} in a modality even when the modality-specific data volume expands, making our method more stable and scalable.

\subsubsection{\textup{\textbf{Impact of Token Amount}}}
\noindent Another intriguing aspect to investigate pertains to the two token amount hyper-parameters $m$ and $n$ that we set in MT. These parameters dictate the number of high-frequency multi-modal tokens retained and processed by CMEE. More tokens correlate with more intricate interactions in the model with an $O((m+n)^2)$-level increase in time efficiency. This is because the time complexity of the Transformer layer used in CMEE is positively related to the quadratic of the sequence length. Therefore, we explored the performance of the MMKGC concerning the variation of the parameters $m$ and $n$.  The experimental results are depicted in Figure \ref{figure::token} as a 3D bar chart. From the figure we can observe that the model performance shows an increasing and then decreasing trend with the increase in the number of tokens $m$ and $n$. This pattern is more distinct with increasing visual tokens, while the text modality experiences minor variations.

\subsubsection{\textup{\textbf{Impact of Different Tokenizers}}}
\begin{figure}[]
  \centering
\includegraphics[width=0.85\columnwidth]{pictures/backbone.pdf}

  \caption{MMKGC results using different tokenizers.}
  \label{figure::tokenizers}
\end{figure}

To further explore the robustness of {\model} over different tokenizers, we conduct the experiments on DB15K with more kinds of visual tokenizers (BEiT \cite{BEIT} and VQGAN \cite{VQ-GAN}) and textual tokenizers (BERT \cite{BERT}, RoBERTa \cite{RoBERTa}, and Llama \cite{llama}).  The MMKGC results with their combinations are presented in Figure \ref{figure::tokenizers}. We can conclude that {\model} is stable and robust across different tokenizers, This indicates that {\model} is a generalizable and universal framework, which can integrate different multi-modal backbones. Besides, we can find that larger backbones like Llama \cite{llama} could bring better performance, demonstrating the possibilities of combining our approach with the latest LLM technology.

\begin{table}[t]
\centering
\setlength{\tabcolsep}{0.7mm}{
\begin{tabular}{cl|cccc}
    \toprule
    \multicolumn{2}{c|}{\textbf{Setting}} & \multicolumn{1}{l}{\textbf{MRR}} & \multicolumn{1}{l}{\textbf{H@10}} & \multicolumn{1}{l}{\textbf{H@3}} & \multicolumn{1}{l}{\textbf{H@1}} \\
    \midrule
     \multirow{4}{*}{\begin{tabular}[c]{@{}c@{}}Model\\ Design\end{tabular}} & w/o MT & 35.48 & 50.89 & 39.09 & 27.48\\
        & w/o Refine& 36.61 & 50.97& 39.89 & 29.13\\
        & w/o CMEE& 34.78 & 50.44& 38.32 & 26.66\\
        & w/o CTE & 34.71 & 50.72& 38.37 & 26.59\\
    \midrule
    \multirow{9}{*}{\begin{tabular}[c]{@{}c@{}}FGCL\end{tabular}} & w/o $\mathcal{L}_{con}$ & 35.99 & 51.31& 39.68 & 27.98\\
        & w/o $e_{sec}$ & 36.82 & 51.75& 40.55 & 29.02\\
        & w/o $\mathbf{s}(e)$ & 37.62 & 52.46& 40.91 & 29.97\\
        & w/o $\mathbf{v}(e)$ & 37.24 & 51.18& 40.70 & 29.58\\
        & w/o $\mathbf{w}(e)$ & 37.64 & 52.16& 41.24 & 29.96\\
        & $\lambda=1$ & 37.43 & 52.03 & 40.75 & 29.83\\
        & $\lambda=0.1$ & 37.48 & 52.16& 41.22 & 29.72\\
        & $\lambda=0.001$ & 36.91 & 52.10& 39.89 & 27.99\\
    \midrule  
    \multicolumn{2}{c|}{Full Model\quad $\lambda=0.01$} & \textbf{37.72} & \textbf{52.21}& \textbf{41.26} & \textbf{30.08} \\
        \bottomrule
\end{tabular}
}
\caption{The ablation study results on DB15K.}
\label{table::ablation}
\end{table}

\subsection{Ablation Study}
\label{section::ablation}
To confirm the effectiveness of each module in {\model}, we further conduct an ablation study from two perspectives:the model design, and the FGCL loss. We removed the corresponding modules across different settings and performed MMKGC experiments. The experimental results are presented in Table \ref{table::ablation}.  According to the first part of the experimental results, all of the core modules we designed in the backbone network, the modality tokenization process, and the filtering process critically influence the final prediction. Besides, the design of FGCL also contributes to the model performance, with a contrastive candidate in $\mathcal{C}(e)$ being essential for achieving the SOTA performance. Meanwhile, we explore the influence of the loss weight $\lambda$ of FGCL. As we tuned $\lambda$ in $\{1, 0.1, 0.01, 0.001\}$, the MMKGC results show an increasing and then decreasing trend and reaches state-of-the-art at $\lambda=0.01$. Overall, we can find that the most pivotal modules affecting the overall performance are CMEE and CTE, which extract fine-grained and contextual entity representations to make a modality-aware triple prediction. The FCGL module further makes more of further boosts model performance based on the backbone.

\subsection{Efficiency Analysis}
\label{sec:efficiency}

\begin{figure}[t]
    \centering
  \includegraphics[width=\columnwidth]{pictures/efficiency.pdf}
    \caption{The efficiency-performance trade-off analysis.}
    \label{figure::efficency}
\end{figure}
To validate the efficiency of {\model}, we conduct an efficiency experiment. As shown in Figure \ref{figure::efficency}, we compare the training efficiency and final prediction of {\model} with different $m$ and $n$ against several recent baselines. We find that {\model} achieves the best performance while maintaining relatively good efficiency, realizing a trade-off between performance and efficiency. Besides, we can observe that the total amount of tokens ($m+n$) has a small effect on efficiency. As $m+n$ increases from 8 to 24, {\model}'s efficiency goes from about 8s to about 13s, which is still acceptable in practice.

\subsection{Token Embedding Visualization}
\label{section::visualization}
\begin{figure}[t]
  \centering
\includegraphics[width=0.8\columnwidth]{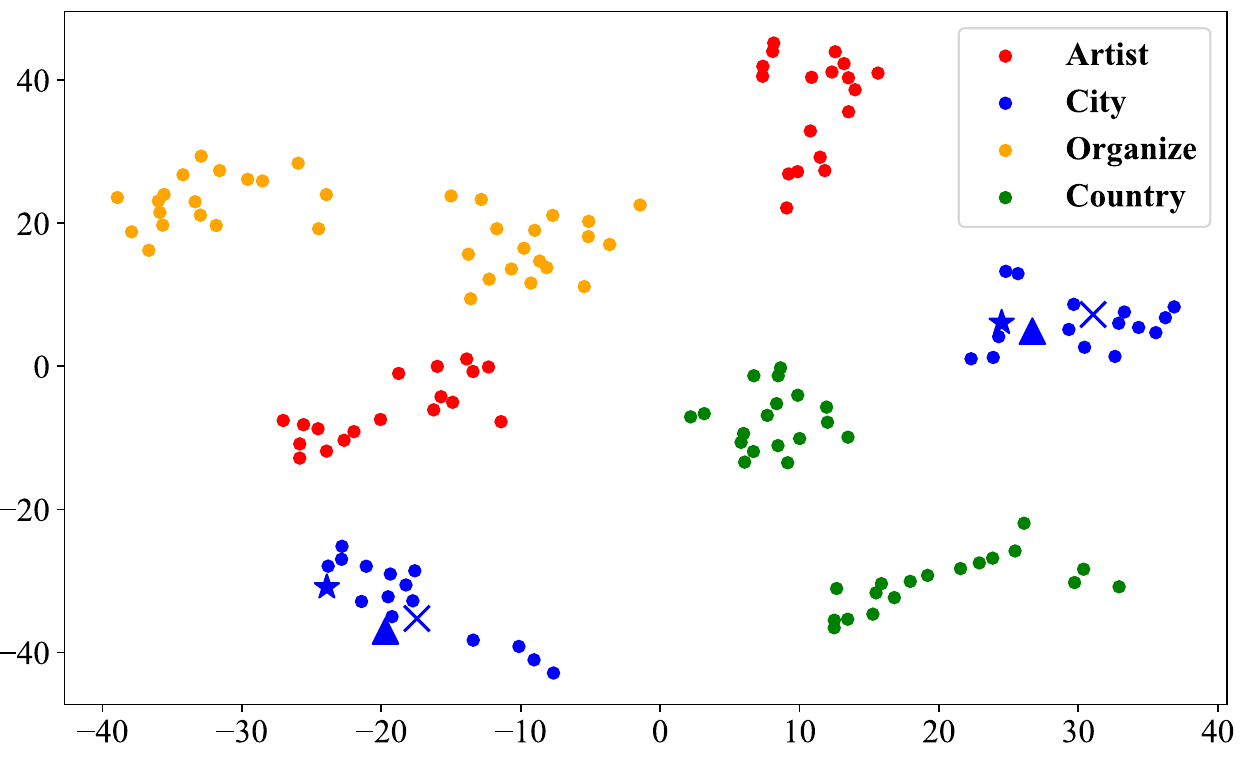}
  \caption{The embedding visualization results of the multi-modal tokens from entities with different categories (Artist, City, Organization, and Country).}
  \label{figure::visualization}
\end{figure}

To give an intuitive view of the learned representations, we conduct an embedding visualization demonstration experiment in this section. We choose four diverse categories of entities (artist, city, organize, country) in DB15K. For each category, we select two entities and perform dimensionality reduction of their multi-modal tokens from the entire input sequence $\mathcal{X}(e)$ employing the t-SNE \cite{tsne} algorithm. For the city category, we uniquely mark the same tokens in both city entities using special marks ($\triangle, \star, \times$) to demonstrate the contextual multi-modal embeddings of the same tokens under different token sequences. Figure \ref{figure::visualization} reveals that each small cluster in the figure represents the tokens of an entity. Furthermore, tokens in entities of the same category can easily form clusters, displaying a certain degree of distinction between tokens from diverse entities. This indicates that the learned token embeddings are distinguishable. Also, identical tokens under differing contexts revealed slight variances, which underscores their potential to provide unique roles for different entities, even if originated from the same token feature. Altogether, this visualization validates the effectiveness of our approach by revealing the distribution of multi-modal tokens.

\section{Conclusion}
In this paper, we focus on the problem of capturing fine-grained semantic information in MMKGs. We propose a new framework {\model} to tokenize, fuse, and augment multi-modal entity representations. Experiments on public benchmarks demonstrate the effectiveness, reliability, reasonableness, and interpretability of our design. In the future, we will focus on processing and interpreting the fine-grained multi-modal information in MMKGs.

\section*{Acknowledgements}
This work is founded by the National Natural Science Foundation of China (NSFCU23B2055 / NSFCU19B2027 / NSFC62306276), Zhejiang Provincial Natural Science Foundation of China (No. LQ23F020017), Yongjiang Talent Introduction Programme (2022A-238-G), and Fundamental Research Funds for the Central Universities (226-2023-00138). This work was supported by AntGroup.

\bibliography{aaai25}

\begin{thebibliography}{52}
\providecommand{\natexlab}[1]{#1}

\bibitem[{Bordes et~al.(2013)Bordes, Usunier, García-Durán, Weston, and Yakhnenko}]{bordes_translating_2013-TransE}
Bordes, A.; Usunier, N.; García-Durán, A.; Weston, J.; and Yakhnenko, O. 2013.
\newblock Translating {Embeddings} for {Modeling} {Multi}-relational {Data}.
\newblock In \emph{{NIPS}}, 2787--2795.

\bibitem[{Cao et~al.(2022)Cao, Xu, Yang, He, Cao, and Huang}]{cao_otkge_2022-OTKGE}
Cao, Z.; Xu, Q.; Yang, Z.; He, Y.; Cao, X.; and Huang, Q. 2022.
\newblock {OTKGE}: {Multi}-modal {Knowledge} {Graph} {Embeddings} via {Optimal} {Transport}.
\newblock In \emph{{NeurIPS}}.

\bibitem[{Chao et~al.(2021)Chao, He, Wang, and Chu}]{chao_pairre_2021-PairRE}
Chao, L.; He, J.; Wang, T.; and Chu, W. 2021.
\newblock {PairRE}: {Knowledge} {Graph} {Embeddings} via {Paired} {Relation} {Vectors}.
\newblock In \emph{Proc. of {ACL}}.

\bibitem[{Chen et~al.(2024)Chen, Zhang, Fang, Geng, Guo, Chen, Li, Zhang, Chen, Zhu, Li, Liu, Pan, Zhang, and Chen}]{MMKG-Survey}
Chen, Z.; Zhang, Y.; Fang, Y.; Geng, Y.; Guo, L.; Chen, X.; Li, Q.; Zhang, W.; Chen, J.; Zhu, Y.; Li, J.; Liu, X.; Pan, J.~Z.; Zhang, N.; and Chen, H. 2024.
\newblock Knowledge Graphs Meet Multi-Modal Learning: {A} Comprehensive Survey.
\newblock \emph{CoRR}, abs/2402.05391.

\bibitem[{Devlin et~al.(2019)Devlin, Chang, Lee, and Toutanova}]{BERT}
Devlin, J.; Chang, M.; Lee, K.; and Toutanova, K. 2019.
\newblock {BERT:} Pre-training of Deep Bidirectional Transformers for Language Understanding.
\newblock In \emph{{NAACL-HLT} {(1)}}, 4171--4186. Association for Computational Linguistics.

\bibitem[{Dong et~al.(2024)Dong, Zhang, Zhou, Zha, Zheng, and Huang}]{MMKG-LLM}
Dong, J.; Zhang, Q.; Zhou, H.; Zha, D.; Zheng, P.; and Huang, X. 2024.
\newblock Modality-Aware Integration with Large Language Models for Knowledge-based Visual Question Answering.
\newblock \emph{CoRR}, abs/2402.12728.

\bibitem[{Esser, Rombach, and Ommer(2021)}]{VQ-GAN}
Esser, P.; Rombach, R.; and Ommer, B. 2021.
\newblock Taming Transformers for High-Resolution Image Synthesis.
\newblock In \emph{{CVPR}}, 12873--12883. Computer Vision Foundation / {IEEE}.

\bibitem[{Gage(1994)}]{BPE}
Gage, P. 1994.
\newblock A new algorithm for data compression.
\newblock \emph{The C Users Journal}, 12(2): 23--38.

\bibitem[{Gao, Yao, and Chen(2021)}]{SIMCSE}
Gao, T.; Yao, X.; and Chen, D. 2021.
\newblock SimCSE: Simple Contrastive Learning of Sentence Embeddings.
\newblock In \emph{{EMNLP} {(1)}}, 6894--6910. Association for Computational Linguistics.

\bibitem[{Han et~al.(2018)Han, Cao, Lv, Lin, Liu, Sun, and Li}]{DBLP:conf/emnlp/OpenKE}
Han, X.; Cao, S.; Lv, X.; Lin, Y.; Liu, Z.; Sun, M.; and Li, J. 2018.
\newblock OpenKE: An Open Toolkit for Knowledge Embedding.
\newblock In \emph{Proc. of EMNLP}.

\bibitem[{Ivana, Carl, and Timothy(2019)}]{Tucker}
Ivana, B.; Carl, A.; and Timothy, M.~H. 2019.
\newblock TuckER: Tensor Factorization for Knowledge Graph Completion.
\newblock In \emph{{EMNLP/IJCNLP} {(1)}}, 5184--5193. Association for Computational Linguistics.

\bibitem[{Ji et~al.(2015)Ji, He, Xu, Liu, and Zhao}]{ji_knowledge_2015-TransD}
Ji, G.; He, S.; Xu, L.; Liu, K.; and Zhao, J. 2015.
\newblock Knowledge {Graph} {Embedding} via {Dynamic} {Mapping} {Matrix}.
\newblock In \emph{{ACL} (1)}, 687--696. The Association for Computer Linguistics.

\bibitem[{Kingma and Ba(2015)}]{DBLP:journals/corr/KingmaB14-Adam}
Kingma, D.~P.; and Ba, J. 2015.
\newblock Adam: {A} Method for Stochastic Optimization.
\newblock In \emph{{ICLR} (Poster)}.

\bibitem[{Kudo(2018)}]{ULM}
Kudo, T. 2018.
\newblock Subword Regularization: Improving Neural Network Translation Models with Multiple Subword Candidates.
\newblock In \emph{{ACL} {(1)}}, 66--75. Association for Computational Linguistics.

\bibitem[{Lee et~al.(2023)Lee, Chung, Lee, Jo, and Whang}]{lee_vista_2023-VISTA}
Lee, J.; Chung, C.; Lee, H.; Jo, S.; and Whang, J.~J. 2023.
\newblock {VISTA}: {Visual}-{Textual} {Knowledge} {Graph} {Representation} {Learning}.
\newblock In \emph{{EMNLP} ({Findings})}, 7314--7328. Association for Computational Linguistics.

\bibitem[{Li et~al.(2023)Li, Zhao, Xu, Zhang, and Xing}]{li_imf_2023-IMF}
Li, X.; Zhao, X.; Xu, J.; Zhang, Y.; and Xing, C. 2023.
\newblock {IMF}: {Interactive} {Multimodal} {Fusion} {Model} for {Link} {Prediction}.
\newblock In \emph{{WWW}}, 2572--2580. ACM.

\bibitem[{Liang et~al.(2024{\natexlab{a}})Liang, Liu, Zhou, Tu, Wen, Yang, Dong, and Liu}]{contrastive}
Liang, K.; Liu, Y.; Zhou, S.; Tu, W.; Wen, Y.; Yang, X.; Dong, X.; and Liu, X. 2024{\natexlab{a}}.
\newblock Knowledge Graph Contrastive Learning Based on Relation-Symmetrical Structure.
\newblock \emph{{IEEE} Trans. Knowl. Data Eng.}, 36(1): 226--238.

\bibitem[{Liang et~al.(2024{\natexlab{b}})Liang, Meng, Liu, Liu, Tu, Wang, Zhou, Liu, Sun, and He}]{KGSurvey}
Liang, K.; Meng, L.; Liu, M.; Liu, Y.; Tu, W.; Wang, S.; Zhou, S.; Liu, X.; Sun, F.; and He, K. 2024{\natexlab{b}}.
\newblock A Survey of Knowledge Graph Reasoning on Graph Types: Static, Dynamic, and Multi-Modal.
\newblock \emph{{IEEE} Trans. Pattern Anal. Mach. Intell.}, 46(12): 9456--9478.

\bibitem[{Liang et~al.(2024{\natexlab{c}})Liang, Meng, Liu, Liu, Wei, Liu, Tu, Wang, Zhou, and Liu}]{SPTformer}
Liang, K.; Meng, L.; Liu, Y.; Liu, M.; Wei, W.; Liu, S.; Tu, W.; Wang, S.; Zhou, S.; and Liu, X. 2024{\natexlab{c}}.
\newblock Simple Yet Effective: Structure Guided Pre-trained Transformer for Multi-modal Knowledge Graph Reasoning.
\newblock In \emph{{ACM} Multimedia}, 1554--1563. {ACM}.

\bibitem[{Liang et~al.(2024{\natexlab{d}})Liang, Meng, Zhou, Tu, Wang, Liu, Liu, Zhao, Dong, and Liu}]{liang2024mines}
Liang, K.; Meng, L.; Zhou, S.; Tu, W.; Wang, S.; Liu, Y.; Liu, M.; Zhao, L.; Dong, X.; and Liu, X. 2024{\natexlab{d}}.
\newblock MINES: Message Intercommunication for Inductive Relation Reasoning over Neighbor-Enhanced Subgraphs.
\newblock In \emph{Proceedings of the AAAI Conference on Artificial Intelligence}, volume~38, 10645--10653.

\bibitem[{Liu et~al.(2019{\natexlab{a}})Liu, Li, Garc{\'{\i}}a{-}Dur{\'{a}}n, Niepert, O{\~{n}}oro{-}Rubio, and Rosenblum}]{MMKG}
Liu, Y.; Li, H.; Garc{\'{\i}}a{-}Dur{\'{a}}n, A.; Niepert, M.; O{\~{n}}oro{-}Rubio, D.; and Rosenblum, D.~S. 2019{\natexlab{a}}.
\newblock {MMKG:} Multi-modal Knowledge Graphs.
\newblock In \emph{{ESWC}}, volume 11503 of \emph{Lecture Notes in Computer Science}, 459--474. Springer.

\bibitem[{Liu et~al.(2019{\natexlab{b}})Liu, Ott, Goyal, Du, Joshi, Chen, Levy, Lewis, Zettlemoyer, and Stoyanov}]{RoBERTa}
Liu, Y.; Ott, M.; Goyal, N.; Du, J.; Joshi, M.; Chen, D.; Levy, O.; Lewis, M.; Zettlemoyer, L.; and Stoyanov, V. 2019{\natexlab{b}}.
\newblock RoBERTa: {A} Robustly Optimized {BERT} Pretraining Approach.
\newblock \emph{CoRR}, abs/1907.11692.

\bibitem[{Lu et~al.(2022)Lu, Wang, Jiang, He, and Liu}]{DBLP:journals/apin/LuWJHL22-MMKRL}
Lu, X.; Wang, L.; Jiang, Z.; He, S.; and Liu, S. 2022.
\newblock {MMKRL:} {A} robust embedding approach for multi-modal knowledge graph representation learning.
\newblock \emph{Appl. Intell.}, 52(7): 7480--7497.

\bibitem[{Paszke et~al.(2019)Paszke, Gross, Massa, Lerer, Bradbury, Chanan, Killeen, Lin, Gimelshein, Antiga, Desmaison, K{\"{o}}pf, Yang, DeVito, Raison, Tejani, Chilamkurthy, Steiner, Fang, Bai, and Chintala}]{pytorch}
Paszke, A.; Gross, S.; Massa, F.; Lerer, A.; Bradbury, J.; Chanan, G.; Killeen, T.; Lin, Z.; Gimelshein, N.; Antiga, L.; Desmaison, A.; K{\"{o}}pf, A.; Yang, E.~Z.; DeVito, Z.; Raison, M.; Tejani, A.; Chilamkurthy, S.; Steiner, B.; Fang, L.; Bai, J.; and Chintala, S. 2019.
\newblock PyTorch: An Imperative Style, High-Performance Deep Learning Library.
\newblock In \emph{NeurIPS}, 8024--8035.

\bibitem[{Peng et~al.(2022)Peng, Dong, Bao, Ye, and Wei}]{BEIT}
Peng, Z.; Dong, L.; Bao, H.; Ye, Q.; and Wei, F. 2022.
\newblock BEiT v2: Masked Image Modeling with Vector-Quantized Visual Tokenizers.
\newblock \emph{CoRR}, abs/2208.06366.

\bibitem[{Ryoo et~al.(2021)Ryoo, Piergiovanni, Arnab, Dehghani, and Angelova}]{VideoToken}
Ryoo, M.~S.; Piergiovanni, A.~J.; Arnab, A.; Dehghani, M.; and Angelova, A. 2021.
\newblock TokenLearner: Adaptive Space-Time Tokenization for Videos.
\newblock In \emph{NeurIPS}, 12786--12797.

\bibitem[{Sergieh et~al.(2018)Sergieh, Botschen, Gurevych, and Roth}]{sergieh_multimodal_2018-TBKGC}
Sergieh, H.~M.; Botschen, T.; Gurevych, I.; and Roth, S. 2018.
\newblock A {Multimodal} {Translation}-{Based} {Approach} for {Knowledge} {Graph} {Representation} {Learning}.
\newblock In \emph{*{SEM}@{NAACL}-{HLT}}, 225--234. Association for Computational Linguistics.

\bibitem[{Simonyan and Zisserman(2015)}]{VGG}
Simonyan, K.; and Zisserman, A. 2015.
\newblock Very Deep Convolutional Networks for Large-Scale Image Recognition.
\newblock In \emph{{ICLR}}.

\bibitem[{Sun et~al.(2020)Sun, Cao, Zhao, Wan, Zhou, Zhang, Wang, and Zheng}]{MMKG-rec}
Sun, R.; Cao, X.; Zhao, Y.; Wan, J.; Zhou, K.; Zhang, F.; Wang, Z.; and Zheng, K. 2020.
\newblock Multi-modal Knowledge Graphs for Recommender Systems.
\newblock In \emph{{CIKM}}, 1405--1414. {ACM}.

\bibitem[{Sun et~al.(2019)Sun, Deng, Nie, and Tang}]{sun_rotate_2019-RotatE}
Sun, Z.; Deng, Z.-H.; Nie, J.-Y.; and Tang, J. 2019.
\newblock {RotatE}: {Knowledge} {Graph} {Embedding} by {Relational} {Rotation} in {Complex} {Space}.
\newblock In \emph{{ICLR} ({Poster})}. OpenReview.net.

\bibitem[{Tang et~al.(2020)Tang, Huang, Wang, He, and Zhou}]{tang_orthogonal_2020-OTE}
Tang, Y.; Huang, J.; Wang, G.; He, X.; and Zhou, B. 2020.
\newblock Orthogonal {Relation} {Transforms} with {Graph} {Context} {Modeling} for {Knowledge} {Graph} {Embedding}.
\newblock In \emph{Proc. of {ACL}}.

\bibitem[{Touvron et~al.(2023)Touvron, Lavril, Izacard, Martinet, Lachaux, Lacroix, Rozi{\`{e}}re, Goyal, Hambro, Azhar, Rodriguez, Joulin, Grave, and Lample}]{llama}
Touvron, H.; Lavril, T.; Izacard, G.; Martinet, X.; Lachaux, M.; Lacroix, T.; Rozi{\`{e}}re, B.; Goyal, N.; Hambro, E.; Azhar, F.; Rodriguez, A.; Joulin, A.; Grave, E.; and Lample, G. 2023.
\newblock LLaMA: Open and Efficient Foundation Language Models.
\newblock \emph{CoRR}, abs/2302.13971.

\bibitem[{Trouillon et~al.(2016)Trouillon, Welbl, Riedel, Gaussier, and Bouchard}]{trouillon_complex_2016-ComplEx}
Trouillon, T.; Welbl, J.; Riedel, S.; Gaussier, {\'{E}}.; and Bouchard, G. 2016.
\newblock Complex Embeddings for Simple Link Prediction.
\newblock In \emph{{ICML}}, volume~48 of \emph{{JMLR} Workshop and Conference Proceedings}, 2071--2080. JMLR.org.

\bibitem[{van~den Oord, Vinyals, and Kavukcuoglu(2017)}]{VQ-VAE}
van~den Oord, A.; Vinyals, O.; and Kavukcuoglu, K. 2017.
\newblock Neural Discrete Representation Learning.
\newblock In \emph{{NIPS}}, 6306--6315.

\bibitem[{Van~der Maaten and Hinton(2008)}]{tsne}
Van~der Maaten, L.; and Hinton, G. 2008.
\newblock Visualizing data using t-SNE.
\newblock \emph{Journal of machine learning research}, 9(11).

\bibitem[{Vaswani et~al.(2017)Vaswani, Shazeer, Parmar, Uszkoreit, Jones, Gomez, Kaiser, and Polosukhin}]{transformer}
Vaswani, A.; Shazeer, N.; Parmar, N.; Uszkoreit, J.; Jones, L.; Gomez, A.~N.; Kaiser, L.; and Polosukhin, I. 2017.
\newblock Attention is All you Need.
\newblock In \emph{{NIPS}}, 5998--6008.

\bibitem[{Wang et~al.(2021)Wang, Wang, Yang, Zhang, Chen, and Qi}]{wang_is_2021-RSME}
Wang, M.; Wang, S.; Yang, H.; Zhang, Z.; Chen, X.; and Qi, G. 2021.
\newblock Is {Visual} {Context} {Really} {Helpful} for {Knowledge} {Graph}? {A} {Representation} {Learning} {Perspective}.
\newblock In \emph{{ACM} {Multimedia}}, 2735--2743. ACM.

\bibitem[{Wang et~al.(2023)Wang, Meng, Chen, Meng, Lv, and Zhu}]{DBLP:conf/mm/WangMCML023-TIVA}
Wang, X.; Meng, B.; Chen, H.; Meng, Y.; Lv, K.; and Zhu, W. 2023.
\newblock {TIVA-KG:} {A} Multimodal Knowledge Graph with Text, Image, Video and Audio.
\newblock In \emph{{ACM} Multimedia}, 2391--2399. {ACM}.

\bibitem[{Wang et~al.(2019)Wang, Li, Li, and Zeng}]{wang_multimodal_2019-TransAE}
Wang, Z.; Li, L.; Li, Q.; and Zeng, D. 2019.
\newblock Multimodal {Data} {Enhanced} {Representation} {Learning} for {Knowledge} {Graphs}.
\newblock In \emph{{IJCNN}}, 1--8. IEEE.

\bibitem[{Wilbur and Sirotkin(1992)}]{stopwords}
Wilbur, W.~J.; and Sirotkin, K. 1992.
\newblock The automatic identification of stop words.
\newblock \emph{J. Inf. Sci.}, 18(1): 45--55.

\bibitem[{Wu et~al.(2016)Wu, Schuster, Chen, Le, Norouzi, Macherey, Krikun, Cao, Gao, Macherey, Klingner, Shah, Johnson, Liu, Kaiser, Gouws, Kato, Kudo, Kazawa, Stevens, Kurian, Patil, Wang, Young, Smith, Riesa, Rudnick, Vinyals, Corrado, Hughes, and Dean}]{WordPiece}
Wu, Y.; Schuster, M.; Chen, Z.; Le, Q.~V.; Norouzi, M.; Macherey, W.; Krikun, M.; Cao, Y.; Gao, Q.; Macherey, K.; Klingner, J.; Shah, A.; Johnson, M.; Liu, X.; Kaiser, L.; Gouws, S.; Kato, Y.; Kudo, T.; Kazawa, H.; Stevens, K.; Kurian, G.; Patil, N.; Wang, W.; Young, C.; Smith, J.; Riesa, J.; Rudnick, A.; Vinyals, O.; Corrado, G.; Hughes, M.; and Dean, J. 2016.
\newblock Google's Neural Machine Translation System: Bridging the Gap between Human and Machine Translation.
\newblock \emph{CoRR}, abs/1609.08144.

\bibitem[{Xie et~al.(2017)Xie, Liu, Luan, and Sun}]{xie_image-embodied_2017-IKRL}
Xie, R.; Liu, Z.; Luan, H.; and Sun, M. 2017.
\newblock Image-embodied {Knowledge} {Representation} {Learning}.
\newblock In \emph{{IJCAI}}, 3140--3146. ijcai.org.

\bibitem[{Xu et~al.(2022)Xu, Xu, Wu, Zhou, and Chen}]{MMRNS}
Xu, D.; Xu, T.; Wu, S.; Zhou, J.; and Chen, E. 2022.
\newblock Relation-enhanced {Negative} {Sampling} for {Multimodal} {Knowledge} {Graph} {Completion}.
\newblock In \emph{{ACM} {Multimedia}}, 3857--3866. ACM.

\bibitem[{Xu et~al.(2023)Xu, Zhou, Xu, Xia, Liu, Chen, and Dou}]{xu_multimodal_2023-CamE}
Xu, D.; Zhou, J.; Xu, T.; Xia, Y.; Liu, J.; Chen, E.; and Dou, D. 2023.
\newblock Multimodal {Biological} {Knowledge} {Graph} {Completion} via {Triple} {Co}-{Attention} {Mechanism}.
\newblock In \emph{{ICDE}}, 3928--3941. IEEE.

\bibitem[{Yang et~al.(2015)Yang, Yih, He, Gao, and Deng}]{yang_embedding_2015-DistMult}
Yang, B.; Yih, W.-t.; He, X.; Gao, J.; and Deng, L. 2015.
\newblock Embedding {Entities} and {Relations} for {Learning} and {Inference} in {Knowledge} {Bases}.
\newblock In \emph{{ICLR} ({Poster})}.

\bibitem[{Yao, Mao, and Luo(2019)}]{DBLP:journals/corr/KG-BERT}
Yao, L.; Mao, C.; and Luo, Y. 2019.
\newblock {KG-BERT:} {BERT} for Knowledge Graph Completion.
\newblock \emph{CoRR}, abs/1909.03193.

\bibitem[{Zhang, Chen, and Zhang(2023{\natexlab{a}})}]{DBLP:conf/ijcnn/ZhangCZ23-MANS}
Zhang, Y.; Chen, M.; and Zhang, W. 2023{\natexlab{a}}.
\newblock Modality-Aware Negative Sampling for Multi-modal Knowledge Graph Embedding.
\newblock In \emph{{IJCNN}}, 1--8. {IEEE}.

\bibitem[{Zhang et~al.(2024)Zhang, Chen, Liang, Chen, and Zhang}]{MAT}
Zhang, Y.; Chen, Z.; Liang, L.; Chen, H.; and Zhang, W. 2024.
\newblock Unleashing the Power of Imbalanced Modality Information for Multi-modal Knowledge Graph Completion.
\newblock \emph{CoRR}, abs/2402.15444.

\bibitem[{Zhang, Chen, and Zhang(2023{\natexlab{b}})}]{MACO}
Zhang, Y.; Chen, Z.; and Zhang, W. 2023{\natexlab{b}}.
\newblock {MACO:} {A} Modality Adversarial and Contrastive Framework for Modality-Missing Multi-modal Knowledge Graph Completion.
\newblock In \emph{{NLPCC} {(1)}}, volume 14302 of \emph{Lecture Notes in Computer Science}, 123--134. Springer.

\bibitem[{Zhang and Zhang(2022)}]{DBLP:journals/corr/abs-2209-07084-VBKGC}
Zhang, Y.; and Zhang, W. 2022.
\newblock Knowledge Graph Completion with Pre-trained Multimodal Transformer and Twins Negative Sampling.
\newblock \emph{CoRR}, abs/2209.07084.

\bibitem[{Zhao et~al.(2022)Zhao, Cai, Wu, Zhang, Zhang, Zhao, and Jiang}]{zhao_mose_2022-MOSE}
Zhao, Y.; Cai, X.; Wu, Y.; Zhang, H.; Zhang, Y.; Zhao, G.; and Jiang, N. 2022.
\newblock {MoSE}: {Modality} {Split} and {Ensemble} for {Multimodal} {Knowledge} {Graph} {Completion}.
\newblock In \emph{{EMNLP}}, 10527--10536. Association for Computational Linguistics.

\bibitem[{Zhu et~al.(2021)Zhu, Zhao, Zhang, Ye, Chen, Zhang, and Chen}]{MMKG-pretraining}
Zhu, Y.; Zhao, H.; Zhang, W.; Ye, G.; Chen, H.; Zhang, N.; and Chen, H. 2021.
\newblock Knowledge Perceived Multi-modal Pretraining in E-commerce.
\newblock In \emph{{ACM} Multimedia}, 2744--2752. {ACM}.

\end{thebibliography}

\clearpage
\appendix
\section{Related Works}
\label{appendix::related_works}
\subsection{Multi-modal Knowledge graph completion (MMKGC)}
\textbf{MMKGs} \cite{MMKG-Survey} are knowledge graphs with rich multi-modal information like images, text descriptions, audio, and videos \cite{DBLP:conf/mm/WangMCML023-TIVA}. Due to the incompleteness of the knowledge graphs, knowledge graph completion (KGC) \cite{bordes_translating_2013-TransE, ji_knowledge_2015-TransD, yang_embedding_2015-DistMult, trouillon_complex_2016-ComplEx, sun_rotate_2019-RotatE} is a popular research topic to automatically discover unobserved knowledge triples by learning from the triple structure. Multi-modal knowledge graph completion (MMKGC) aims to predict missing triples in the given MMKGs collaboratively leveraging the extra multi-modal information from entities.

\par Existing MMKGC methods mainly make new improvements in three perspectives: (1) multi-modal fusion and interaction, (2) integrated decision, and (3) negative sampling. Methods of the first category \cite{xie_image-embodied_2017-IKRL, sergieh_multimodal_2018-TBKGC, wang_is_2021-RSME, cao_otkge_2022-OTKGE, xu_multimodal_2023-CamE} design complex mechanisms to achieve multi-modal fusion and interaction in the representation space. For example, OTKGE proposes an optimal transport-based multi-modal fusion strategy to find the optimal weights for multi-modal fusion. The second category methods \cite{zhao_mose_2022-MOSE, li_imf_2023-IMF} usually learn a discriminate model for each modality and ensemble them to make joint decisions. IMF \cite{li_imf_2023-IMF} proposes an interactive multi-modal fusion method to achieve multi-modal fusion and learns four different MMKGC models with different modality information to achieve joint decisions. The third category methods \cite{MMRNS, DBLP:conf/ijcnn/ZhangCZ23-MANS, MACO, MAT} aim to enhance the negative sampling process \cite{bordes_translating_2013-TransE} with the multi-modal information of entities to generate high-quality negative samples. Overall, these MMKGC methods usually leverage the multi-modal information by extracting feature representations from pre-trained models \cite{VGG, BERT}. However, the feature processing them neglects the fine-grained semantic information in each modality. We will solve this problem by tokenizing the modality information into fine-grained tokens.

\subsection{Multi-modal Information Tokenization}

\par Tokenization is a widely used technology in the NLP field to process the input text into a token sequence and learn fine-grained textual representations of strings and subwords. Due to the characteristics of textual modality itself, tokenization is very effective and has been widely used in language models (LM). For example, BPE \cite{BPE}, WordPiece \cite{WordPiece}, and ULM \cite{ULM} are the most famous tokenization methods. 
For information in other modalities, tokenization becomes relatively difficult as there is no clear separation point for these modalities, which is different from texts. vector quantization (VQ) \cite{VQ-VAE, VQ-GAN} is an important technology proposed to map large-scale data into a fixed-length discrete codebook, where each code in the codebook is a vector representing certain specific features. Therefore, the non-textual modality information can be firstly processed into patch sequences and then each patch is mapped into a discrete code, which can be regarded as multi-modal tokens and further leveraged in many tasks \cite{BEIT, VideoToken}. VQ has the advantage of compressing multimodal data while preserving a wide variety of fine-grained modal features with discrete code. For example, BEIT-v2 \cite{BEIT} processes each image into 196 patches in the size of 16x16 and maps each patch into a discrete code. In our work, we will also employ VQ and tokenization to process the multi-modal information in MMKGs and obtain fine-grained multi-modal representation for entities.

\section{Methodology}
\subsection{Refinement Details}
\label{appendix::refinement}
During the tokenization process, it's common to encounter duplicate tokens since certain subwords might appear multiple times within a sentence, and analogous semantic elements could recur across entity images. Therefore, we count the occurrence frequency of each token, retaining a predetermined quantity of the most common tokens for each modality. Additionally, we remove the stop words \cite{stopwords} in the textual descriptions as their contribution to the entity semantics is minimal. After such refinement, we reserve $m$ visual tokens and $n$ textual tokens for each entity in the MMKG. For those entities with insufficient tokens to scant or absent modality data, we add a special padding token to fill the gaps.

\par Unlike existing MMKGC methods, the MT technique converts the modality information into more fine-grained discrete tokens. When facing multiple information in one modality (e.g. multiple images for one entity), traditional MMKGCs would make an aggregation (e.g. mean averaging \cite{MAT}) on them before. However, MT preserves a sequence of tokens that represent the most prevalent features from various raw data sources, which can be more stable and scalable for increasing modality information. We will demonstrate this point in the experiments.
\section{Experiments}
\subsection{Evaluation Protocol and Metrics}
\label{appendix::evaluation}
\par In the inference stage, MMKGC models with predict the missing entities for a given query $(h, r, ?)$ or $(?, r, t)$. Taking tail prediction $(h, r, ?)$ for instance, MMKGC models will treat each entity $e\in\mathcal{E}$ as a candidate entity and calculate its corresponding score $(h, r, e)$. Further, the models are evaluated by the rank of the golden answer $(h, r, t)$ against all candidates, which means rank-based metrics \cite{bordes_translating_2013-TransE, sun_rotate_2019-RotatE} will be employed for performance evaluation. It is mirrored for the head predictions and the overall performance usually considers both head and tail predictions on the test triples.

\par Therefore, we apply rank-based metrics like MRR and Hit@K as the evaluation metrics. The final results are the average of both head prediction and tail prediction. MRR and Hit@K can be denoted as:

\begin{small}
    \begin{equation}
       \mathbf{MRR}=\frac{1}{|\mathcal{T}_{test}|}\sum_{i=1}^{|\mathcal{T}_{test}|}(\frac{1}{r_{h,i}}+\frac{1}{r_{t,i}})
    \end{equation}
    \begin{equation}
       \mathbf{Hit@K}=\frac{1}{|\mathcal{T}_{test}|}\sum_{i=1}^{|\mathcal{T}_{test}|}(\mathbf{1}(r_{h,i} \leq K)+\mathbf{1}(r_{t,i} \leq K))
    \end{equation}
\end{small}
where $r_{h, i}$ and $r_{t, i}$ are the results of head prediction and tail prediction,  and $\mathcal{T}_{test}$ is the test triple set.

\subsection{Detailed Baselines}
\label{appendix::baselines}
We focus on the ability of the methods to exploit multi-modal information in our comparison and select three types of baselines:

\par (1). \textbf{Uni-modal KGC methods}: TransE \cite{bordes_translating_2013-TransE}, DistMult \cite{yang_embedding_2015-DistMult}, ComplEx \cite{trouillon_complex_2016-ComplEx}, RotatE \cite{sun_rotate_2019-RotatE}, PairRE \cite{chao_pairre_2021-PairRE}, GC-OTE \cite{tang_orthogonal_2020-OTE}, and Tucker \cite{Tucker}. These conventional methods only consider the structure information in their model design.

\par (2). \textbf{MMKGC methods}: IKRL \cite{xie_image-embodied_2017-IKRL}, TBKGC \cite{sergieh_multimodal_2018-TBKGC}, TransAE \cite{wang_multimodal_2019-TransAE}, MMKRL \cite{DBLP:journals/apin/LuWJHL22-MMKRL}, RSME \cite{wang_is_2021-RSME}, VBKGC \cite{DBLP:journals/corr/abs-2209-07084-VBKGC}, OTKGE \cite{cao_otkge_2022-OTKGE}, MACO \cite{MACO}, IMF \cite{li_imf_2023-IMF}, QEB \cite{DBLP:conf/mm/WangMCML023-TIVA}, VISTA \cite{lee_vista_2023-VISTA}, and AdaMF \cite{MAT}. These methods consider image and textual information in the MMKGC models and achieve multi-modal fusion with different settings.

\par (3). \textbf{Negative sampling methods}: MANS \cite{DBLP:conf/ijcnn/ZhangCZ23-MANS}, and MMRNS \cite{MMRNS}. These methods utilize the multi-modal information in the MMKGs to generate high-quality negative samples for training.

\par Some of the methods like KG-BERT \cite{DBLP:journals/corr/KG-BERT} for fine-tuning pre-trained language models are orthogonal to our design, so we did not compare our method to them.

\subsection{Implementation Details}
\label{appendix::implementation}
The visual tokenizer we used is from BEiT, which has a codebook of 8192 different tokens, each with a feature dimension of 32. the textual tokenizer is from BERT, with a total of 32,000 tokens of 768 dimensions. In fact, due to the coverage of the dataset itself, not all tokens in the vocabulary and codebook are used. Specifically, the coverages of our visual and textual tokens are around 40\% and 35\% respectively. We also tried other visual and textual tokenizers like VQGAN \cite{VQ-GAN}, Llama \cite{llama}, and RoBERTa \cite{RoBERTa} in the further exploration.

\par During training, we keep 3 images for each entity with visual information in the main experiments. The feature dimensions of the visual and textual tokens are 32 and 768 designed by the original models. During training, we set the training epoch to 2000, the batch size to 1024, and the embedding dimension to 256. For the transformer layers, we employ 1 transformer layer for both CMEE and CTE with 4 attention heads. We set the dropout layer with $p\in\{0.3, 0.4, 0,5\}$. The max token number $m$ and $n$ are tuned in $\{4, 8, 12\}$ and the contrastive loss weight $\lambda$ is tuned in $\{1, 0.1, 0.01, 0.001\}$. The contrastive temperature $\tau$ is tuned in $\{0.1, 0.5, 1.0\}$. We optimize the model with Adam \cite{DBLP:journals/corr/KingmaB14-Adam} optimizer and the learning rate is searched in $\{1e^{-3}, 1e^{-4}, 5e^{-4}\}$. For the baseline models, we reuse their official code or reproduce their results based on OpenKE \cite{DBLP:conf/emnlp/OpenKE}. 
\par All the experiments are conducted on a Linux server with one NVIDIA A800 GPU, taking 1 to 5 hours to finish the training and evaluation on different datasets. Though using transformer encoders, our method is sufficiently efficient, with each epoch requiring 11s (DB15K) on a single GPU. This is closer to the traditional MMKGC method and faster than some of the baselines of recent years (e.g., MMRNS 23s, OTKGE 64s). Besides, the training can be performed on consumer-level GPUs. A single 24G 3090 is enough. The best hyper-parameters to achieve SOTA performance shown in the paper is presented in Table \ref{table::param}.

\begin{table}[h]
\caption{Addtional results on DB15K to explore the model performance with different hyper-parameter settings.}

\label{table::param}
\centering
\resizebox{\columnwidth}{!}{
\begin{tabular}{c|ccc}
\toprule
Parameter  & DB15K    & MKG-W & MKG-Y \\ \midrule
embedding dimension & 256 & 256 & 256  \\
dropout & 0.3 & 0.4 & 0.4 \\
temperature $\tau$  & 0.5 & 0.5 & 0.5 \\
loss weight $\lambda$ & 0.01 & 0.001 & 0.001 \\
learning rate & $1e^{-3}$ & $5e^{-4}$  & $5e^{-4}$\\
\bottomrule
\end{tabular}
}
\end{table}

\subsection{Hyper-parameter Analysis}
\label{appendix::hyper-parameter}
our approach is \textbf{little influenced by the batch size parameter}. The increase in the number of transformer layers may cause the model to overfit, and there will be no significant improvement but slower efficiency. We added \textbf{additional experiments in Table} \ref{table1}

\begin{table}[h]
\caption{Addtional results on DB15K to explore the model performance with different hyper-parameter settings.}

\label{table1}
\centering
\resizebox{\columnwidth}{!}{
\begin{tabular}{cc|cccccc}
\toprule
batch size & transformer layer  & MRR    & Hit@1   & Hit@3 & Hit@10 &\\ \midrule
1024 & 1 & 37.72 & 30.08 & 41.26 & 52.21  \\
512 & 1 & 37.16 & 29.28 & 40.86  & 52.18 \\
2048 & 1 & 37.13 & 29.29 & 40.72  & 52.33 \\
1024 & 2 & 37.37 & 29.67 & 40.77  & 51.84 \\
1024 & 4 & 36.47 & 28.59 & 40.32  & 51.71 \\
\bottomrule
\end{tabular}
}
\end{table}

\end{document}